\documentclass[twoside]{article}

\usepackage{microtype}
\usepackage{graphicx}
\usepackage{subfigure}
\usepackage{booktabs}

\usepackage{hyperref}

\usepackage{amsmath}
\usepackage{amsthm}
\usepackage{amssymb}
\usepackage{booktabs}
\usepackage{paralist}
\usepackage[shortlabels, inline]{enumitem}
\usepackage[np, autolanguage]{numprint}
\usepackage{multirow}
\usepackage{multicol}
\usepackage{xcolor}
\usepackage{relsize}
\usepackage{dsfont}   
\usepackage{xargs}   
\usepackage{eurosym}
\usepackage{acronym}
\usepackage{breqn}
\usepackage{algorithm}
\usepackage{algorithmic}

\acrodef{ML}{machine learning}
\acrodef{GDPR}{General Data Protection Regulation}
\usepackage[skip=0pt]{caption}
\DeclareRobustCommand{\officialeuro}{%
  \ifmmode\expandafter\text\fi
  {\fontencoding{U}\fontfamily{eurosym}\selectfont e}}

\theoremstyle{definition}

\theoremstyle{remark}

\DeclareMathOperator{\softmax}{softmax}

\newcommand{\graph}{\mathcal{G}}
\newcommand{\compgraph}{\mathcal{G}_v}
\newcommand{\compadj}{A_v}
\newcommand{\compdeg}{D_v}
\newcommand{\compnode}{X_v}

\newcommand{\loss}{\mathcal{L}}
\newcommand{\losspred}{\mathcal{L}_{pred}}

\newcommand{\lossdist}{\mathcal{L}_{dist}}
\newcommand{\perturbb}{P}
\newcommand{\perturbl}{\hat{P}}

\newcommand{\OurMethod}{\textsc{CF-GNNExplainer}}
\newcommand{\OurShort}{\textsc{CF-GNN}}
\newcommand{\synone}{\textsc{ba-shapes}}
\newcommand{\synfour}{\textsc{tree-cycles}}
\newcommand{\synfive}{\textsc{tree-grid}}
\newcommand{\gnnexplainer}{\textsc{GNNExplainer}}
\newcommand{\gnnexpshort}{\textsc{GNNExp}}

\newcommand{\cgraph}{subgraph neighborhood}
\newcommand{\baserand}{\textsc{random}}
\newcommand{\basekeep}{\textsc{1hop}}
\newcommand{\baserm}{\textsc{rm-1hop}}

\newcounter{todocnt}

\newlength\myindent
\newcommand\bindent{%
  \begingroup
  \setlength{\itemindent}{\myindent}
  \addtolength{\algorithmicindent}{\myindent}
}
\newcommand\eindent{\endgroup}

\usepackage[accepted]{aistats2022}
\usepackage[round]{natbib}

\begin{document}

%

%
\runningauthor{Lucic, ter Hoeve, Tolomei, de Rijke, Silvestri}

\twocolumn[

\aistatstitle{\OurMethod{}: Counterfactual Explanations for Graph Neural Networks}


\aistatsauthor{ Ana Lucic $^1$  \quad  Maartje ter Hoeve $^1$ \quad  Gabriele Tolomei $^2$ \quad Maarten de Rijke $^1$ \quad  Fabrizio Silvestri $^2$ }

\aistatsaddress{ $^{1}$ University of Amsterdam \And  $^{2}$ Sapienza University of Rome} ]

\begin{abstract}
Given the increasing promise of graph neural networks (GNNs) in real-world applications, several methods have been developed for explaining their predictions. Existing methods for interpreting predictions from GNNs have primarily focused on generating subgraphs that are especially relevant for a particular prediction. 
However, such methods are not counterfactual (CF) in nature: given a prediction, we want to understand how the prediction can be changed in order to achieve an alternative outcome. 
In this work, we propose a method for generating CF explanations for GNNs: the minimal perturbation to the input (graph) data such that the prediction changes. 
Using only edge deletions, we find that our method, \OurMethod{}, can generate CF explanations for the majority of instances across three widely used datasets for GNN explanations, while removing less than 3 edges on average, with at least 94\% accuracy. 
This indicates that \OurMethod{} primarily removes edges that are crucial for the original predictions, resulting in minimal CF explanations. 

\end{abstract}


\section{\uppercase{Introduction}}
\label{section:introduction}
Advances in machine learning (ML) have led to breakthroughs in several areas of science and engineering,  ranging from computer vision, to natural language processing, to conversational assistants. 
Parallel to the increased performance of ML systems, there is an increasing call for the ``understandability'' of ML models ~\citep{goebel-2018-explainable}. 
Understanding \emph{why} an ML model returns a certain output in response to a given input is important for a variety of reasons such as model debugging, aiding decison-making, or fulfilling legal requirements \citep{gdpr}. 
Having certified methods for interpreting ML predictions will help enable their use across a variety of applications~\citep{miller-2017-explanations}.

Explainable AI (XAI) refers to the set of techniques ``\textit{focused on exposing complex AI models to humans in a systematic and interpretable manner}''~\citep{samekexplainable}. A large body of work on XAI has emerged in recent years~\citep{guidotti-2018-survey,bodria2021benchmarking}. Counterfactual (CF) explanations are used to explain predictions of individual instances in the form: ``If X had been different, Y would not have occurred''~\citep{stepin2021survey,karimi_model-agnostic_2019,schut_generating_2021}. 
CF explanations are based on CF examples: modified versions of the input sample that result in an alternative output (i.e., prediction). 
If the proposed modifications are also \emph{actionable}, this is referred to as achieving recourse \citep{ustun_actionable_2019,karimi2020survey}. 

To motivate our problem, consider an ML application for computational biology: drug discovery is a task that involves generating new molecules that can be used for medicinal purposes \citep{stokes_deep_2020,xie2021mars}. 
Given a candidate molecule, a GNN can predict if this molecule has a certain property that would make it effective in treating a particular disease \citep{wieder_compact_2020,guo2021fewshot,nguyen2020metalearning}.
If the GNN predicts it does not have this desirable property, CF explanations can help identify the minimal change required such that the molecule is predicted to have this property. 
This could help not only inform the design of a new molecule that has this property, but also understand the molecular structures that contribute to this property.

Although GNNs have shown state-of-the-art results on tasks involving graph data \citep{zitnik_modeling_2018,deac_drug-drug_2019}, existing methods for explaining the predictions of GNNs have primarily focused on generating subgraphs that are relevant for a particular prediction~\citep{yuan2020explainability,baldassarre_explainability_2019,duval2021graphsvx,lin_causal_2021,luo_parameterized_2020,pope_explainability_2019,schlichtkrull_interpreting_2020,vu2020pgmexplainer,ying_gnnexplainer_2019,yuan_subgraph_2021}. 
However, \emph{none of these methods are able to identify the minimal subgraph automatically} -- they all require the user to specify the size of the subgraph, $S$, in advance. 
We show that even if we adapt existing methods to the CF explanation problem, and try varying values for $S$, such methods are not able to produce valid, accurate CF explanations, and are therefore not well-suited to solve the CF explanation problem. 
To address this gap, we propose \OurMethod{}, a method for generating CF explanations for GNNs. 

Similar to other CF methods for tabular or image data proposed in the literature~\citep{verma2020counterfactual, karimi2020survey}, \OurMethod{} works by perturbing input data at the instance-level. 
Unlike previous methods, \OurMethod{} can generate CF explanations for graph data. 
In particular, our method iteratively removes edges from the original adjacency matrix based on matrix sparsification techniques, keeping track of the perturbation that leads to a change in prediction, and returning the perturbation with the smallest change w.r.t.\ the number of edges. 

\begin{figure}[h]
    \centering
    \includegraphics[width=\columnwidth]{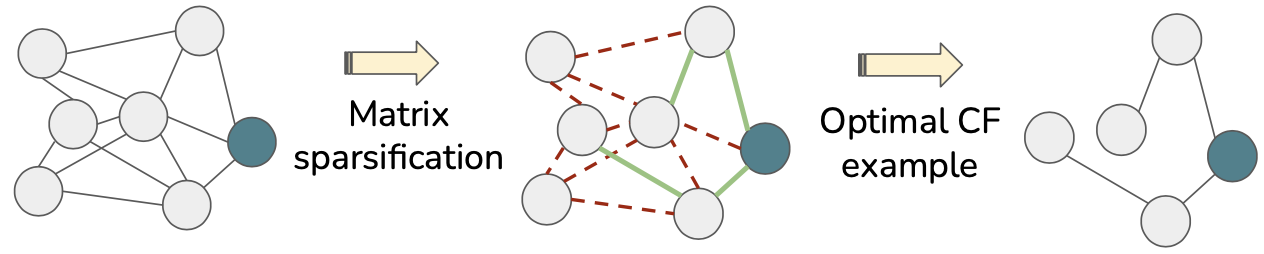}
    \caption{Intuition of counterfactual example generation by \OurMethod{}.}
    \label{fig:visual}
\end{figure}

We evaluate \OurMethod{} on three public datasets for GNN explanations and measure its effectiveness using four metrics: fidelity, explanation size, sparsity, and accuracy. We find that \OurMethod{} is able to generate CF examples with at least 94\% accuracy, while removing fewer than 3 edges on average. 
We make the following contributions:
\begin{itemize}[leftmargin=*,nosep]
    \item We formalize the problem of generating CF explanations for GNNs (Section~\ref{section:problem-formulation}). 
    \item We propose \OurMethod{}, a novel method for explaining predictions from GNNs (Section~\ref{section:method}). 
    \item We propose an experimental setup for holistically evaluating CF explanations for GNNs (Section~\ref{section:experimental-setup}).
\end{itemize}   

The implementation of \OurMethod{} is available at \url{https://github.com/cf-gnnexplainer}.


\section{\uppercase{Related Work}}
\label{section:relatedwork}

Since our work is a counterfactual XAI approach for GNNs, it is related to GNN explainability (Section~\ref{section:rw-gnnxai}) as well as counterfactual explanations (Section~\ref{section:rw-cfexp}). It is also related to adversarial attack methods (Section~\ref{section:rw-adversarial}).

\subsection{GNN Explainability}
\label{section:rw-gnnxai}
Several GNN XAI approaches have been proposed -- a recent survey of the most relevant work is presented by~\citet{yuan2020explainability}.
However, unlike our work, {\em none} of the methods in this survey generate CF explanations. 

The majority of existing GNN XAI methods provide an explanation in the form of a subgraph of the original graph that is deemed to be important for the prediction \citep{yuan2020explainability,baldassarre_explainability_2019,duval2021graphsvx,lin_causal_2021,luo_parameterized_2020,pope_explainability_2019,schlichtkrull_interpreting_2020,vu2020pgmexplainer,ying_gnnexplainer_2019,yuan_subgraph_2021}. We refer to these as \emph{subgraph-generating methods}. 
Such methods are analogous to popular XAI methods such as LIME \citep{ribeiro-2016-should} or SHAP \citep{lundberg_unified_2017}, which identify relevant features for a particular prediction for tabular, image, or text data. 
All of these methods require the user to specify the size of the explanation, $S$, in advance: the number of features (or edges) to keep. 
In contrast, \OurMethod{} generates CF explanations, which can find the size of the explanation without requiring input from the user. 
Although both types of techniques are meant for explaining GNN predictions, they are solving fundamentally different problems: CF explanations generate the minimal perturbation such that the prediction changes, while subgraph-retrieving methods identify a relevant (and not necessarily minimal) subgraph that matches the original prediction. 

\citet{kang_explaine_2019} also generate CF examples for GNNs, but their work focuses on a different task: link prediction. 
Other GNN XAI methods identify important node features~\citep{huang_graphlime_2020} or similar examples~\citep{faber_contrastive_2020}. 
\citet{yuan_xgnn_2020} and \citet{schnake_xai_2020} generate model-level (i.e., global) explanations for GNNs, which differs from our work since we produce instance-level (i.e., local) explanations. 

\subsection{Counterfactual Explanations}
\label{section:rw-cfexp}
There exists a substantial body of work on CF explanations for tabular, image, and text data \citep{verma2020counterfactual,karimi2020survey,stepin2021survey}.
Some methods treat the underlying classification model as a black-box \citep{laugel_inverse_2017, guidotti_local_2018, lucic2019does}, whereas others make use of the model's inner workings \citep{tolomei_interpretable_2017, wachter_counterfactual_2017, ustun_actionable_2019, kanamori_dace_2020, lucic2020focus}.
All of these methods are based on perturbing feature values to generate CF examples -- they are not equipped to handle graph data with relationships (i.e., edges) between instances (i.e., nodes). In contrast, \OurMethod{} provides CF examples specifically for graph data.

\subsection{Adversarial Attacks}
\label{section:rw-adversarial}
CF examples are also related to adversarial attacks \citep{sun2020adversarial}: they both represent instances obtained from minimal perturbations to the input, which induce changes in the prediction made by the learned model. 
One difference between the two is in the intent: adversarial examples are meant to fool the model, while CF examples are meant to explain the prediction \citep{freiesleben_intriguing_2021,lucic2020focus}. 
In the context of graph data, adversarial attack methods typically make minimal perturbations to the \emph{overall graph} with the intention of degrading overall model performance, as opposed to attacking individual nodes. 
In contrast, we are interested in generating CF examples for individual nodes, as opposed to identifying perturbations to the overall graph. 
We confirm that the CF examples produced by \OurMethod{} are informative and not adversarial by measuring the accuracy of our method (see Section~\ref{section:metrics}).


\section{\uppercase{Background}}
\label{section:background}
In this section, we provide background information on GNNs (Section~\ref{section:gnns-general}) and matrix sparsification (Section~\ref{section:matrix-sparsification}), both of which are necessary for understanding \OurMethod{}. 

\subsection{Graph Neural Networks}
\label{section:gnns-general}
Graphs are structures that represent a set of entities (nodes) and their relations (edges). 
GNNs operate on graphs to produce representations that can be used in downstream tasks such as graph or node classification. The latter is the focus of this work.
We refer to the survey papers by \citet{battaglia_relational_2018} and \citet{chami2021machine} for an overview of existing GNN methods. 

Let $f(A, X; W) \rightarrow y$ be any GNN, where $y$ is the set of possible predicted classes, $A$ is an $n \times n$ adjacency matrix, $X$ is an $n \times p$ feature matrix, and $W$ is the learned weight matrix of $f$. 
In other words, $A$ and $X$ are the inputs of $f$, and $f$ is parameterized by $W$. 

A node's representation is learned by iteratively updating the node's features based on its neighbors' features.   
The number of layers in $f$ determines which neighbors are included: if there are $\ell$ layers, then the node's final representation only includes neighbors that are at most $\ell$ hops away from that node in the graph $\graph$. 
The rest of the nodes in $\graph$ are not relevant for the computation of the node's final representation.  
We define the \emph{\cgraph{}} of a node $v$ as the set of the nodes and edges relevant for the computation of $f(v)$ (i.e., those in the $\ell$-hop neighborhood of $f$), represented as a tuple: $\compgraph = (\compadj, \compnode)$, where $\compadj$ is the subgraph adjacency matrix and $X_v$ is the node feature matrix for nodes that are at most $\ell$ hops away from $v$. We then define a node $v$ as a tuple of the form $v = (\compadj, x)$, where $x$ is the feature vector for $v$. 

\subsection{Matrix Sparsification}
\label{section:matrix-sparsification}
\OurMethod{} uses matrix sparsification to generate CF examples, inspired by~\citet{srinivas_training_2016}, who propose a method for training sparse neural networks. 
Given a weight matrix $W$, a binary sparsification matrix is learned which is multiplied element-wise with $W$ such that some of the entries in $W$ are zeroed out. 
In the work by \cite{srinivas_training_2016}, the objective is to remove entries in the weight matrix in order to reduce the number of parameters in the model. 
In our case, we want to \emph{zero out entries in the adjacency matrix} (i.e., remove edges) in order to generate CF explanations for GNNs. 
That is, we want to remove the important edges -- those that are crucial for the prediction. 

\section{\uppercase{Problem Formulation}}
\label{section:problem-formulation}
In general, a CF example $\bar{x}$ for an instance $x$ according to a trained classifier $f$ is found by perturbing the features of $x$ such that $f(x) \neq f(\bar{x})$ \citep{wachter_counterfactual_2017}. 
An optimal CF example $\bar{x}^*$ is one that minimizes the distance between the original instance and the CF example, according to some distance function $d$, and the resulting optimal CF explanation is $\Delta^*_{x} = \bar{x}^* - x$ \citep{lucic2020focus}. 

For graph data, it may not be enough to simply perturb node features, especially since they are not always available. 
This is why we are interested in generating CF examples by perturbing the graph structure instead. 
In other words, we want to change the relationships between instances (i..e, nodes), rather than change the instances themselves. 
Therefore, a CF example for graph data has the form $\bar{v} = (\bar{\compadj}, x)$, where $x$ is the feature vector and $\bar{\compadj}$ is a perturbed version of  $\compadj$, the adjacency matrix of the subgraph neighborhood of a node $v$. $\bar{\compadj}$ is obtained by removing some edges from $\compadj$, such that $f(v) \neq f(\bar{v})$. 
Following \citet{wachter_counterfactual_2017} and \citet{lucic2020focus}, we generate CF examples by minimizing a loss function of the form:
\begin{align}
\label{eq:loss-graph}
    \mathcal{L} = \losspred(v, \bar{v} \mid f, g) + \beta \lossdist(v, \bar{v} \mid d),
\end{align}
where $v$ is the original node, $f$ is the original model, $g$ is the CF model that generates $\bar{v}$, and $\losspred$ is a prediction loss that encourages $f(v) \neq f(\bar{v})$. 
$\lossdist$ is a distance loss that encourages $\bar{v}$ to be close to $v$, and $\beta$ controls how important $\lossdist$ is compared to $\losspred$. 
We want to find $\bar{v}^*$ that minimizes Eq.~\ref{eq:loss-graph}: this is the optimal CF example for $v$.


\section{\uppercase{Method: \OurMethod{}}}
\label{section:method}

To solve the problem defined in Section~\ref{section:problem-formulation}, we propose \OurMethod{}, which generates $\bar{v} = (\bar{\compadj}, x)$ given a node $v = (\compadj, x)$.
Our method can operate on any GNN model $f$. 
To illustrate our method and avoid cluttered notation, let $f$ be a standard, one-layer Graph Convolutional Network~\citep{kipf_semi_supervised_2017} for node classification:
\begin{align}
    \label{eq:gcn}
    f(A, X; W) = \softmax\left[\tilde{D}^{-1/2} \tilde{A} \tilde{D}^{-1/2} X W \right], 
\end{align}
where $\tilde{A} = A + I$, $I$ is the identity matrix, $\tilde{D}_{ii} = \sum_j \tilde{A}_{ij}$ are entries in the degree matrix $\tilde{D}$, $X$ is the node feature matrix, and $W$ is the weight matrix \citep{kipf_semi_supervised_2017}.

\subsection{Adjacency Matrix Perturbation}
\label{section:adjacency-perturb}
First, we define $\bar{\compadj} = \perturbb \odot \compadj$, where $\perturbb$ is a binary perturbation matrix that sparsifies $\compadj$. 
Our aim is to find $\perturbb$ for a given node $v$ such that $f(\compadj, x) \neq f(\perturbb \odot \compadj, x$). 
To find $\perturbb$, we build upon the method by \citet{srinivas_training_2016} for training sparse neural networks (see Section~\ref{section:background}), where our objective is to zero out entries in the adjacency matrix (i.e., remove edges).
That is, we want to find $\perturbb$ that minimally perturbs $\compadj$, and use it to compute $\bar{\compadj} = \perturbb \odot \compadj$. 
If an element $\perturbb_{{i,j}} = 0$, this results in the deletion of the edge between node $i$ and node $j$. 
When $\perturbb$ is a matrix of ones, this indicates that all edges in $\compadj$ are used in the forward pass. 

Similar to the work by \citet{srinivas_training_2016}, we first generate an intermediate, real-valued matrix $\perturbl$ with entries in $\left[0, 1\right]$, apply a sigmoid transformation, then threshold the entries to arrive at a binary $\perturbb$: entries greater than or equal to 0.5 become 1, while those below 0.5 become 0. 
In the case of undirected graphs (i.e., those with symmetric adjacency matrices), we first generate a perturbation vector which we then use to populate $\perturbl$ in a symmetric manner, instead of generating $\perturbl$ directly.

\subsection{Counterfactual Generating Model}
We want our perturbation matrix $\perturbb$ to only act on $\compadj$, not $\tilde{\compadj}$, in order to preserve self-loops in the message passing of $f$. This is because we always want a node representation update to include its own representation from the previous layer. 
Therefore we first rewrite Eq.~\ref{eq:gcn} for our illustrative one-layer case to isolate $\compadj$: 
\begin{equation}
\begin{split}
    &\mbox{}\hspace*{-2mm}f(\compadj, \compnode; W) = \\ 
    &\mbox{}\hspace*{-2mm}\softmax\left[(\compdeg + I)^{-1/2} (\compadj + I) (\compdeg + I)^{-1/2} \compnode W\right]
    \hspace*{-3mm}\mbox{}
\end{split}    
    \label{eq:gcn3}
\end{equation}
To generate CFs, we propose a new function $g$, which is based on $f$, but it is parameterized by $\perturbb$ instead of $W$. 
We update the degree matrix $\compdeg$ based on $\perturbb \odot \compadj$, add the identity matrix to account for self-loops (as in $\tilde{\compdeg}$ in Eq.~\ref{eq:gcn}), and call this $\bar{\compdeg}$: 
\begin{equation}
    \begin{split}
    \label{eq:cf}
    &g(\compadj, \compnode, W; \perturbb) = \\
    &\softmax\left[\bar{\compdeg}^{-1/2} (\perturbb \odot \compadj + I) \bar{\compdeg}^{-1/2} \compnode W\right]
    \end{split}
\end{equation}
In other words, $f$ learns the weight matrix while holding the data constant, while $g$ 
generates new data points (i.e., CF examples) while holding the weight matrix (i.e., model) constant. 
Another distinction between $f$ and $g$ is that the aim of $f$ is to find the optimal set of weights that generalizes well on an unseen test set, while the objective of $g$ is to generate an optimal CF example, given a particular node (i.e., $\bar{v}$ is the output of $g$).

\subsection{Loss Function Optimization}
We generate $\perturbb$ by minimizing Eq.~\ref{eq:loss-graph}, adopting the negative log-likelihood (NLL) loss  for $\losspred$:
\begin{equation}
\begin{split}
    &\losspred(v, \bar{v}|f, g) ={}\\
    &- \mathds{1}\left[f(v) = f(\bar{v})\right] \cdot \mathcal{L}_{NLL}(f(v), g(\bar{v})).
\end{split}    
    \label{eq:loss-pred}
\end{equation}
Since we do not want $f(\bar{v})$ to match $f(v)$, we put a negative sign in front of $\losspred$ and include an indicator function to ensure the loss is active as long as $f(\bar{v}) = f(v)$. 
Note that $f$ and $g$ have the same weight matrix $W$ -- the main difference is that $g$ also includes the perturbation matrix $\perturbb$.

$\lossdist$ can be based on any differentiable distance function. 
In our case, we take $d$ to be the element-wise difference between $v$ and $\bar{v}$, corresponding to the difference between $\compadj$ and $\bar{\compadj}$: the number of edges removed. 
For undirected graphs, we divide this value by 2 to account for the symmetry in the adjacency matrices. 
When updating $\perturbb$, we take the gradient of Eq.~\ref{eq:loss-graph} with respect to the intermediate $\perturbl$, \emph{not} the binary $\perturbb$.

\begin{algorithm*}[t]
    \caption{\OurMethod{}: given a node $v = (\compadj, x)$ where $f(v) = y$, generate the minimal perturbation, $\bar{v} = (\bar{\compadj}, x)$, such that $f(\bar{v}) \neq y$.}
    
    \label{alg:cf-gnnexplainer}
    \begin{multicols}{2}
    \begin{algorithmic}
        \STATE {\bfseries Input:} node $v = (\compadj, x)$, trained GNN model $f$, CF model $g$, loss function $\loss$, learning rate $\alpha$, number of iterations $K$, distance function $d$. 
        
        \STATE
        
        \STATE $f(v) = y$ \bindent \textcolor{gray}{\textit{\# Get GNN prediction} } \eindent
        


        \STATE $\perturbl \gets J_n$ \bindent  \textcolor{gray}{\textit{\# Initialization} } \eindent
        
        \STATE $\bar{v}^* = \left[ \:\right]$
        
        \STATE
        
        \FOR{$K$ iterations} \label{line:start_loop}
            \STATE $\bar{v}$ = \textsc{get\_cf\_example()}
            \STATE $\loss \gets \loss(v, \bar{v}, f, g)$ \bindent \textcolor{gray}{\textit{\# Eq~\ref{eq:loss-graph} \& Eq~\ref{eq:loss-pred}} } \eindent
            \STATE $\perturbl \gets \perturbl + \alpha \nabla_{\perturbl} \loss$ \bindent \textcolor{gray}{\textit{\# Update $\perturbl$}} \eindent
        \ENDFOR
        
        \STATE
        \end{algorithmic}
        \columnbreak

        \begin{algorithmic}
        \STATE \textbf{Function} \textsc{get\_cf\_example()}
        
        \bindent
        
        \STATE $\perturbb \gets \text{threshold}(\sigma(\perturbl))$
        
        \STATE $\bar{\compadj} \gets \perturbb \odot \compadj$
        \STATE $\bar{v}_{cand} \gets (\bar{\compadj}, x)$
        
        \IF{$f(v) \neq f(\bar{v}_{cand})$} 
            \STATE $\bar{v} \gets \bar{v}_{cand}$
            
            \IF{not $\bar{v}^*$}
            	\STATE $\bar{v}^* \gets \bar{v}$ \bindent \textcolor{gray}{\textit{\# First CF}} \eindent
            
            \ELSIF{$d(v, \bar{v}) \leq d(v, \bar{v}^*)$}
            
                \STATE $\bar{v}^* \gets \bar{v}$ \bindent \textcolor{gray}{\textit{\# Keep track of best CF}} \eindent
            \ENDIF
                    
        \ENDIF
        
        \RETURN{$\bar{v}^*$}
        
        \eindent
        
        
        
    \end{algorithmic}
    \end{multicols}
\vspace{-5mm}
\end{algorithm*}

\subsection{\OurMethod{} }

We call our method \OurMethod{} and summarize its details in Algorithm~\ref{alg:cf-gnnexplainer}. 
Given an node in the test set $v$, we first obtain its original prediction from $f$ and initialize ${\perturbl}$ as a matrix of ones, $J_n$, to initially retain all edges. 
Next, we run \OurMethod{} for $K$ iterations.  
To find a CF example, we use Eq.~\ref{eq:cf}. 

First, we compute $\perturbb$ by thresholding $\perturbl$ (see Section~\ref{section:adjacency-perturb}). 
Then we use $\perturbb$ to obtain the sparsified adjacency matrix that gives us a candidate CF example, $\bar{v}_{cand}$. 
This example is then fed to the original GNN, $f$, and if $f$ predicts a different output than for the original node, we have found a valid CF example, $\bar{v}$. 
We keep track of the ``best'' CF example (i.e., the most minimal according to $d$), and return this as the optimal CF example $\bar{v}^*$ after $K$ iterations. 
Between iterations, we compute the loss following Eqn~\ref{eq:loss-graph} and \ref{eq:loss-pred}, and update $\perturbl$ based on the gradient of the loss. 
In the end, we retrieve the optimal CF explanation $\Delta_v^* = v - \bar{v}^* $. 

\subsection{Complexity}
\label{section:complexity}
\OurMethod{} has time complexity $O(KN^2)$, where $N$ is the number of nodes in the subgraph neighbourhood and $K$ is the number of iterations. We note that high complexity is common for local XAI methods (i.e., SHAP, GNNExplainer, etc), but in practice, one typically only generates explanations for a subset of the dataset.


\section{\uppercase{Experimental Setup}}
\label{section:experimental-setup}

In this section, we outline our experimental setup for evaluating \OurMethod{}, including the datasets and models used (Section~\ref{section:datasets}), the baselines we compare against (Section~\ref{section:baselines}), the evaluation metrics (Section~\ref{section:metrics}), and the hyperparameter search method (Section~\ref{section:hyperparams}). 
In total, we run approximately 375 hours of experiments on one Nvidia TitanX Pascal GPU with access to 12GB RAM.

\subsection{Datasets and Models}
\label{section:datasets}
Given the challenges associated with defining and evaluating the accuracy of XAI methods~\citep{doshi-2017-towards}, we first focus on synthetic tasks where we know the ground-truth explanations. 
Although there exist real graph classification datasets with ground-truth explanations~\citep{mutag_dataset}, there do not exist any real node classification datasets with ground-truth explanations, which is the task we focus on in this paper. 
Building such a dataset would be an excellent contribution, but is outside the scope of this paper.

In our experiments, we use the \synfour{}, \textsc{tree-grids}, \synone{} datasets from \citet{ying_gnnexplainer_2019}. 
These datasets were created specifically for the task of explaining node classification predictions from GNNs. 
Each dataset consists of (i) a base graph, (ii) motifs that are attached to random nodes of the base graph, and (iii) additional edges that are randomly added to the overall graph. 
They are all undirected graphs. 
The classification task is to determine whether or not the nodes are part of the motif. 
The purpose of these datasets is to have a ground-truth for the ``correctness'' of an explanation: for nodes in the motifs, the explanation is the motif itself \citep{luo_parameterized_2020}. 
The dataset statistics are available in Table 1.

\synfour{} consists of a binary tree base graph with 6-cycle motifs, \textsc{tree-grids} also has a binary tree as its base graph, but with 3$\times$3 grids as the motifs. 
For \synone{}, the base graph is a Barabasi-Albert (BA) graph with house-shaped motifs, where each motif consists of 5 nodes (one for the top of the house, two in the middle, and two on the bottom). 
Here, there are four possible classes (not in motif, in motif: top, middle, bottom). 
We note that compared to the other two datasets, the \synone{} dataset is much more densely connected -- the node degree is more than twice as high as that of the \synfour{} or \synfive{} datasets, and the average number of nodes and edges in each node's computation graph is order(s) of magnitude larger. 
We use the same experimental setup (i.e., dataset splits, model architecture) as \citet{ying_gnnexplainer_2019} to train a 3-layer GCN (hidden size = 20) for each task. 
Our GCNs have at least 87\% accuracy on the test set.

\begin{table}[]
\caption{Dataset statistics. The \# edges in the motif indicates the size of the ground truth (GT) explanation. }
\label{table:stats}
\centering
\begin{tabular}{lrrr}
\toprule
                          & \multicolumn{1}{c}{\textsc{Tree}}   & \multicolumn{1}{c}{\textsc{Tree}} & \multicolumn{1}{c}{\textsc{BA}}     \\
                          & \multicolumn{1}{c}{\textsc{Cycles}} & \multicolumn{1}{c}{\textsc{Grid}} & \multicolumn{1}{c}{\textsc{Shapes}} \\ 
\midrule
\# classes                 & 2                         & 2               & 4                          \\ 

\# nodes in motif                 & 6                        & 9            & 5                        \\
\# edges in motif      (GT)            & 6                       & 12             & 6                \\
\midrule
\# nodes in total                  & 871                        & 1231            & 700                        \\
\# edges in total                  & 1950                       & 3410             & 4100                \\

\midrule
Avg node degree           & 2.27                       & 2.77                     & 5.87                       \\
Avg \# nodes in $\compadj$ & 19.12                      & 30.69                    & 304.40                     \\
Avg \# edges in $\compadj$ & 18.99                      & 33.94                    & 1106.24                    \\
\bottomrule
\end{tabular}
\end{table}

\subsection{Baselines}
\label{section:baselines}
Since existing GNN XAI methods give explanations in the form of relevant subgraphs as opposed to CF examples, it is not straightforward to identify baselines for our experiments that ensure a fair comparison between methods. 
To evaluate \OurMethod{}, we compare against 4 baselines: \baserand{}, \basekeep{}, \baserm{}, and \gnnexplainer{}.
The random perturbation is meant as a sanity check. 
We randomly initialize the entries of $\perturbl \in \left[-1, 1\right]$ and apply the same sigmoid transformation and thresholding as described in Section~\ref{section:adjacency-perturb}. 
We repeat this $K$ times and keep track of the most minimal perturbation resulting in a CF example. 
Next, we compare against baselines that are based on the ego graph of $v$ (i.e., its 1-hop neighbourhood): \basekeep{} keeps all edges in the ego graph of $v$, while \baserm{} removes all edges in the ego graph of $v$. 

Our fourth baseline is based on \gnnexplainer{} by \citet{ying_gnnexplainer_2019}, which identifies the $S$ most relevant edges for the prediction (i.e., the most relevant subgraph of size $S$). 
To generate CF explanations, we remove the subgraph generated by \gnnexplainer{}. 
We include this method in our experiments in order to have a baseline based on a prominent GNN XAI method, but we note that subgraph-retrieving methods like \gnnexplainer{} are not meant for generating CF explanations. 
Unlike our method, \gnnexplainer{} cannot automatically find a \emph{minimal} subgraph and therefore requires the user to determine the number of edges to keep in advance (i.e., the value of $S$). 
As a result, we cannot evaluate how minimal its CF explanations are, but we can compare it against our method in terms of 
\begin{inparaenum}[(i)]
	\item its ability to generate valid CF examples, and 
	\item how accurate those CF examples are.
\end{inparaenum}
We report results on \gnnexplainer{} for $ S \in \{1, 2, 3, 4, 5,$ GT$\}$, where GT is the size of the ground truth explanation (i.e., the number of edges in the motif, see Table~\ref{table:stats}).

\subsection{Metrics}
\label{section:metrics}
We generate a CF example for each node in the graph separately and evaluate in terms of four metrics.

\textbf{Fidelity:} is defined as the proportion of nodes where the original predictions match the prediction for the explanations \citep{molnar2019,ribeiro-2016-should}. Since we generate CF examples, we do not want the original prediction to match the prediction for the explanation, so we want a low value for fidelity. 

\textbf{Explanation Size:} is the number of removed edges. It corresponds to the $\lossdist$ term in Equation~\ref{eq:loss-graph}: the difference between the original $\compadj$ and the counterfactual $\bar{\compadj}$. Since we want to have \emph{minimal} explanations, we want a small value for this metric. Note that we cannot evaluate this metric for \gnnexplainer{}. 

\textbf{Sparsity:} measures the proportion of edges in $\compadj$ that are removed \citep{yuan2020explainability}. A value of 0 indicates all edges in $\compadj$ were removed. Since we want \emph{minimal} explanations, we want a value close to 1. Note that we cannot evaluate this metric for \gnnexplainer{}.

\textbf{Accuracy:} is the mean proportion of explanations that are ``correct''. Following \citet{ying_gnnexplainer_2019} and \citet{luo_parameterized_2020}, we only compute accuracy for nodes that are originally predicted as being part of the motifs, since accuracy can only be computed on instances for which we know the ground truth explanations. 
Given that we want \emph{minimal} explanations, we consider an explanation to be correct if it \emph{exclusively} involves edges that are inside the motifs (i.e., only removes edges that are within the motifs). 

The exact calculations of all metrics can be found in the public code base at \url{https://github.com/cf-gnnexplainer}.

\begin{table*}[]
\centering
\caption{Results comparing our method to \baserand{}, \basekeep{}, and \baserm{}. Below each metric, $\blacktriangledown$ indicates a low value is desirable, while $\blacktriangle$ indicates a high value is desirable.}
\label{table:results1}
\setlength{\tabcolsep}{4pt}
\begin{tabular}{lrrrr rrrr rrrr}
\toprule
\multicolumn{1}{c}{} & \multicolumn{4}{c}{\synfour{}}                                                                                                                 & \multicolumn{4}{c}{\synfive{}}                                                                                                                   & \multicolumn{4}{c}{\synone{}}                                                                                                                  \\ 
\cmidrule(r){2-5}\cmidrule(r){6-9}\cmidrule{10-13} 
               & \multicolumn{1}{c}{\textit{Fid.}} & \multicolumn{1}{c}{\textit{Size}} & \multicolumn{1}{c}{\textit{Spars.}} & \multicolumn{1}{c}{\textit{Acc.}} & \multicolumn{1}{c}{\textit{Fid.}} & \multicolumn{1}{c}{\textit{Size}} & \multicolumn{1}{c}{\textit{Spars.}} & \multicolumn{1}{c}{\textit{Acc.}} & \multicolumn{1}{c}{\textit{Fid.}} & \multicolumn{1}{c}{\textit{Size}} & \multicolumn{1}{c}{\textit{Spars.}} & \multicolumn{1}{c}{\textit{Acc.}} \\


Method & \multicolumn{1}{c}{$\blacktriangledown$} &\multicolumn{1}{c}{$\blacktriangledown$} &\multicolumn{1}{c}{$\blacktriangle$} & \multicolumn{1}{c}{$\blacktriangle$} & \multicolumn{1}{c}{$\blacktriangledown$} &\multicolumn{1}{c}{$\blacktriangledown$} &\multicolumn{1}{c}{$\blacktriangle$} & \multicolumn{1}{c}{$\blacktriangle$} & \multicolumn{1}{c}{$\blacktriangledown$} &\multicolumn{1}{c}{$\blacktriangledown$} &\multicolumn{1}{c}{$\blacktriangle$} & \multicolumn{1}{c}{$\blacktriangle$} \\
\midrule
\baserand{}               & \textbf{0.00}                     & 4.70                              & 0.79                                & 0.63                               & \textbf{0.00}                     & 9.06                              & 0.75                                & 0.77                               & \textbf{0.00}                     & 503.31                            & 0.58                                & 0.17                              \\
\basekeep{}                 & 0.32                              & 15.64                             & 0.13                                & 0.45                               & 0.32                              & 29.30                             & 0.09                                & 0.72                               & 0.60                              & 504.18                            & 0.05                                & 0.18                              \\
\baserm{}              & 0.46                              & 2.11                              & 0.89                                & ---                                  & 0.61                              & 2.27                              & 0.92                                & ---                                  & 0.21                              & 10.56                             & 0.97                                & \textbf{0.99}                     \\


\midrule
\OurMethod{}               & 0.21                              & \textbf{2.09}                     & \textbf{0.90}                       & \textbf{0.94}                      & 0.07                              & \textbf{1.47}                     & \textbf{0.94}                       & \textbf{0.96}                      & 0.39                              & \textbf{2.39}                     & \textbf{0.99}                       & 0.96                 \\
\bottomrule
\end{tabular}
\end{table*}

\subsection{Hyperparameter Search}
\label{section:hyperparams}
We experiment with different optimizers and hyperparameter values for the number of iterations $K$, the trade-off parameter $\beta$, the learning rate $\alpha$, and the Nesterov momentum $m$ (when applicable). 
We choose the setting that produces the most CF examples. 
We test the number of iterations $K \in \{100, 300, 500\}$, the trade-off parameter $\beta \in \{0.1, 0.5\}$, the learning rate $\alpha \in \{0.005, 0.01, 0.1, 1\}$, and the Nesterov momentum $m \in \{0, 0.5, 0.7, 0.9\}$. 
We test Adam, SGD and AdaDelta as optimizers. 
We find that for all three datasets, the SGD optimizer gives the best results, with $k = 500$, $\beta = 0.5$, and $\alpha = 0.1$. 
For the \synfour{} and \synfive{} datasets, we set $m = 0$, while for the \synone{} dataset, we use $m = 0.9$.


\section{\uppercase{Results}}
\label{section:results}

We evaluate \OurMethod{} in terms of the metrics outlined in Section~\ref{section:metrics}. 
The results are shown in Table~\ref{table:results1} and Table~\ref{table:results-gnnexplainer}.  
In almost all settings, we find that \OurMethod{} outperforms the baselines in terms of explanation size, sparsity, and accuracy, which shows that \OurMethod{} satisfies our objective of finding accurate, minimal CF examples. 
In cases where the baselines outperform \OurMethod{} on a particular metric, they perform poorly on the rest of the metrics, or on other datasets.

\subsection{Main Findings}
\textbf{Fidelity:}
 \OurMethod{} outperforms \basekeep{} across all three datasets, and outperforms \baserm{} for \synfour{} and \synfive{} in terms of fidelity. 
We find that \baserand{} has the lowest fidelity in all cases -- it is able to find CF examples for every single node. 
In the following subsections, we will see that this corresponds to poor performance on the other metrics.

\begin{figure*}[]

    \centering

    \includegraphics[scale=0.39]{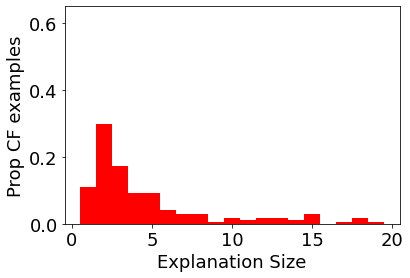}
    \includegraphics[scale=0.39]{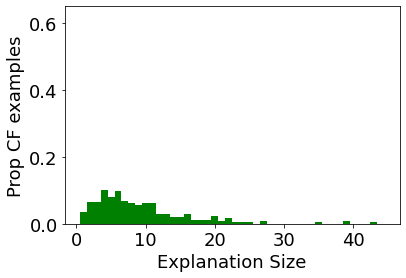}
    \includegraphics[scale=0.39]{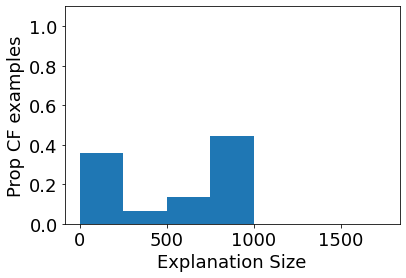}
    
        \caption{Histograms showing the proportion of CF examples that have a certain explanation size from \baserand{}. Note the $x$-axis for \synone{} goes up to 1500. Left: \synfour{}, Middle: \synfive{}, Right: \synone{}.  }
        \label{fig:random-explanation-size}
        
    \includegraphics[scale=0.39]{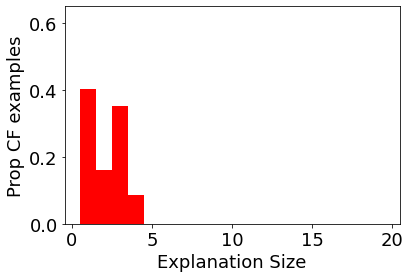}
    \includegraphics[scale=0.39]{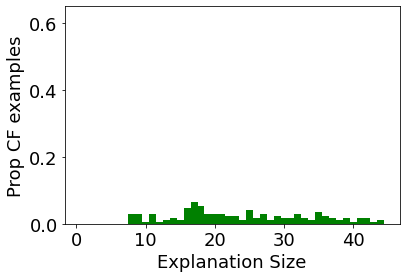}
    \includegraphics[scale=0.39]{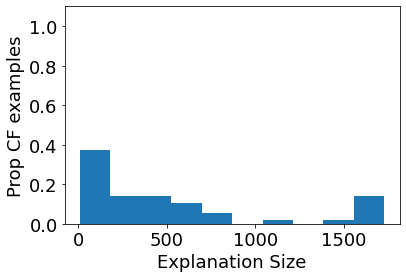}
    
        \caption{Histograms showing the proportion of CF examples that have a certain explanation size from \basekeep{}. Note the $x$-axis for \synone{} goes up to 1500. Left: \synfour{}, Middle: \synfive{}, Right: \synone{}. }
        \label{fig:keep-explanation-size}

    \includegraphics[scale=0.39]{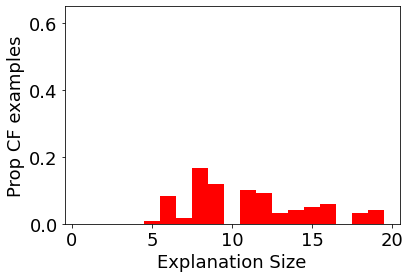}
    \includegraphics[scale=0.39]{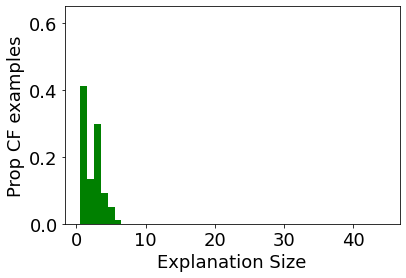}
    \includegraphics[scale=0.39]{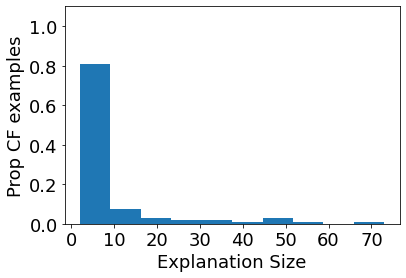}
    
        \caption{Histograms showing the proportion of CF examples that have a certain explanation size from \baserm{}. Note the $x$-axis for \synone{} goes up to 70. Left: \synfour{}, Middle: \synfive{}, Right: \synone{}. }
        \label{fig:remove-explanation-size}
        
    

    \includegraphics[scale=0.38]{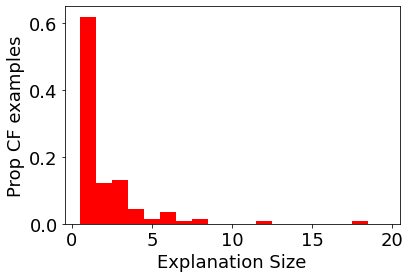}
    \includegraphics[scale=0.38]{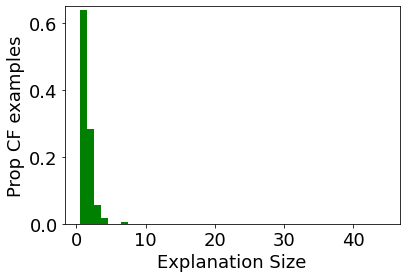}
    \includegraphics[scale=0.38]{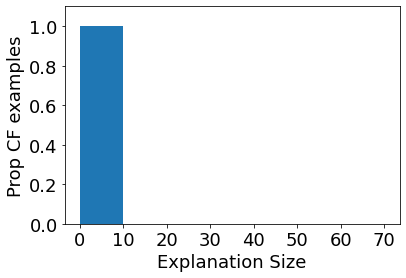}
    
        \caption{Histograms showing the proportion of CF examples that have a certain explanation size from \OurMethod{}. Note the $x$-axis for \synone{} goes up to 70. Left: \synfour{}, Middle: \synfive{}, Right: \synone{}. }
        \label{fig:explanation-size}
        
\end{figure*}

\textbf{Explanation Size:}
Figures~\ref{fig:random-explanation-size} to~\ref{fig:explanation-size} show histograms of the explanation size for \OurMethod{} and the baselines. 

We see that across all three datasets, \OurMethod{} has the smallest (i.e., most minimal) explanation sizes. 
This is especially true when comparing to \baserand{} and \basekeep{} for the \synone{} dataset, where we had to use a different scale for the $x$-axis due to how different the explanation sizes were. 
We postulate that this difference could be because \synone{} is a much more densely connected graph;
it has fewer nodes but more edges compared to the other two datasets, and the average number of nodes and edges in the \cgraph{} is order(s) of magnitude larger (see Table~\ref{table:stats}). 
Therefore, when performing random perturbations, there is substantial opportunity to remove edges that do not necessarily need to be removed, leading to much larger explanation sizes.
When there are many edges in the \cgraph{}, removing everything except the 1-hop neighbourhood, as is done in \basekeep{}, also results in large explanation sizes. 
In contrast, the loss function used by \OurMethod{} ensures that only a few edges are removed, which is the desirable behavior since we want minimal explanations.

\textbf{Sparsity:}
\OurMethod{} outperforms the \baserand{}, \baserm{}, \basekeep{} baselines for all three datasets in terms of sparsity.
We note \OurMethod{} and \baserm{} perform much better on this metric in comparison to the other methods, which aligns with the results from explanation size. 

\textbf{Accuracy:}
We observe that \OurMethod{} has the highest accuracy for the \synfour{} and \synfive{} datasets, whereas \baserm{} has the highest accuracy for \synone{}. 
However, we are unable to calculate the accuracy of \baserm{} for the other two datasets since it is unable to generate \emph{any} CF examples for motif nodes, contributing to the low sparsity on those datasets. 
We observe accuracy levels upwards of 94\% for \OurMethod{} across \emph{all} datasets, indicating that it is consistent in correctly removing edges that are crucial for the initial predictions in the vast majority of cases (see Table~\ref{table:results1}).

\begin{table*}[]
\centering
\caption{Results comparing our method to \gnnexplainer{}. \gnnexplainer{} cannot find $S$ automatically, so we try varying values of $S$. GT indicates the size of the ground truth explanation for each dataset. \OurMethod{} finds $S$ automatically. Below each metric, $\blacktriangledown$ indicates a low value is desirable, while $\blacktriangle$ indicates a high value is desirable.}
\label{table:results-gnnexplainer}
\setlength{\tabcolsep}{4pt}
\begin{tabular}{lrrrr rrrr rrrr}
\toprule
\multicolumn{1}{c}{} & \multicolumn{4}{c}{\synfour{}}                                                                                                                 & \multicolumn{4}{c}{\synfive{}}                                                                                                                   & \multicolumn{4}{c}{\synone{}}                                                                                                                  \\ 
\cmidrule(r){2-5}\cmidrule(r){6-9}\cmidrule{10-13} 
              & \multicolumn{1}{c}{\textit{Fid.}} & \multicolumn{1}{c}{\textit{Size}} & \multicolumn{1}{c}{\textit{Spars.}} & \multicolumn{1}{c}{\textit{Acc.}} & \multicolumn{1}{c}{\textit{Fid.}} & \multicolumn{1}{c}{\textit{Size}} & \multicolumn{1}{c}{\textit{Spars.}} & \multicolumn{1}{c}{\textit{Acc.}} & \multicolumn{1}{c}{\textit{Fid.}} & \multicolumn{1}{c}{\textit{Size}} & \multicolumn{1}{c}{\textit{Spars.}} & \multicolumn{1}{c}{\textit{Acc.}} \\
Method & \multicolumn{1}{c}{$\blacktriangledown$} &\multicolumn{1}{c}{$\blacktriangledown$} &\multicolumn{1}{c}{$\blacktriangle$} & \multicolumn{1}{c}{$\blacktriangle$} & \multicolumn{1}{c}{$\blacktriangledown$} &\multicolumn{1}{c}{$\blacktriangledown$} &\multicolumn{1}{c}{$\blacktriangle$} & \multicolumn{1}{c}{$\blacktriangle$} & \multicolumn{1}{c}{$\blacktriangledown$} &\multicolumn{1}{c}{$\blacktriangledown$} &\multicolumn{1}{c}{$\blacktriangle$} & \multicolumn{1}{c}{$\blacktriangle$} \\
\midrule

\gnnexpshort{} ($S=1$) & 0.65 & 1.00 & 0.92 & 0.61 & 0.69 & 1.00 & 0.96 & 0.79 & 0.90 & 1.00 & 0.94 & 0.52 \\
\gnnexpshort{} ($S=2$) & 0.59 & 2.00 & 0.85 & 0.54 & 0.51 & 2.00 & 0.92 & 0.78 & 0.85 & 2.00 & 0.91 & 0.40  \\
\gnnexpshort{} ($S=3$) & 0.56 & 3.00 & 0.79 & 0.51 & 0.46 & 3.00 & 0.88 & 0.79 & 0.83 & 3.00 & 0.87 & 0.34 \\
\gnnexpshort{} ($S=4$) & 0.58 & 4.00 & 0.72 & 0.48 & 0.42 & 4.00 & 0.84 & 0.79 & 0.83 & 4.00 & 0.83 & 0.31 \\
\gnnexpshort{} ($S=5$) & 0.57 & 5.00 & 0.66 & 0.46 & 0.40 & 5.00 & 0.80  & 0.79 & 0.81 & 5.00 & 0.81 & 0.27 \\
\gnnexpshort{} ($S=$ GT) &  0.55 &	6.00 & 0.57 &	0.46 &	0.35 &	11.83 &	0.53 &	0.74 &	0.82 &	6.00 &	0.79 &	0.24    \\

\midrule
\OurMethod{}               & \textbf{0.21}                              & 2.09                     & 0.90                       & \textbf{0.94}                      & \textbf{0.07}                              & 1.47                     & 0.94                       & \textbf{0.96}                      & \textbf{0.39}                              & 2.39                     & 0.99                       & \textbf{0.96}                 \\
\bottomrule
\end{tabular}
\end{table*}

\subsection{Comparison to \gnnexplainer{}}
Table~\ref{table:results-gnnexplainer} shows the results comparing our method to \gnnexplainer{}. We find that our method outperforms \gnnexplainer{} across all three datasets in terms of both fidelity and accuracy, for all tested values of $S$. 
However, this is not surprising since \gnnexplainer{} is not meant for generating CF explanations, so we cannot expect it to perform well on a task it was not designed for. 
We cannot compare explanation size or sparsity fairly since \gnnexplainer{} requires the user to input the value of $S$.

\subsection{Summary of results} 
Evaluating on four distinct metrics for each dataset gives us a more holistic view of the results. 
We find that across all three datasets, \OurMethod{} can generate CF examples for the majority of nodes in the test set (i.e., low fidelity), while only removing a small number of edges (i.e., low explanation size, high sparsity). For nodes where we know the ground truth (i.e., those in the motifs) we achieve at least 94\% accuracy. 

Although \baserand{} can generate CF examples for every node, they are not very minimal or accurate. 
The latter is also true for \basekeep{} -- in general, it has the worst scores for explanation size, sparsity and accuracy. 
\gnnexplainer{} performs at a similar level as \basekeep{}, indicating that although it is a prominent GNN XAI method, it is not well-suited for solving the CF explanation problem. 

\baserm{} is competitive in terms of explanation size, but it performs poorly in terms of fidelity for the \synfour{} and \synfive{} datasets, and its accuracy on these datasets is unknown since it is unable to generate \emph{any} CF examples for nodes in the motifs. 
These results show that our method is simple and extremely effective in solving the CF explanation task, unlike the baselines we test.


\section{\uppercase{Societal Impact}}
\label{section:limitations}

Researchers have raised concerns about the hidden assumptions behind the use of CF examples~\citep{barocas_hidden_2019}, as well as potentials for misuse~\citep{kasirzadeh2021use}. 
When explaining ML systems through CF examples, it is crucial to account for the context in which the system is deployed. 
CF explanations are not a guarantee to achieving recourse~\citep{ustun_actionable_2019} -- changes suggested should be seen as candidate changes, not absolute solutions, since what is pragmatically actionable differs depending on the context. 

We believe it is crucial for the ML community to invest in developing more rigorous evaluation protocols for XAI methods. 
We suggest that researchers in XAI collaborate with researchers in human-computer interaction to design human-centered user studies about evaluating the utility of XAI methods in practice. 
We are glad to see initiatives for such collaborations already taking place~\citep{ehsan_operationalizing_2020}.

\section{\uppercase{Conclusion}}
\label{section:conclusion}
In this work, we propose \OurMethod{}, a method for generating CF explanations for any GNN. Our simple and effective method is able to generate CF explanations that are (i) minimal, both in terms of the absolute number of edges removed (explanation size), as well as the proportion of the \cgraph{} that is perturbed (sparsity), and (ii) accurate, in terms of removing edges that we know to be crucial for the initial predictions. 

We evaluate our method on three commonly used datasets for GNN explanation tasks and find that these results hold across all three datasets. 
We find that existing GNN XAI methods are not well-suited to solving the CF explanation task, while \OurMethod{} is able to reliably produce minimal, accurate CF explanations. 

In its current form, \OurMethod{} is limited to performing edge deletions in the context of node classification tasks. 
For future work, we plan to incorporate node feature perturbations in our framework and extend \OurMethod{} to accommodate graph classification tasks. 
We also plan to investigate adapting graph attack methods for generating CF explanations, as well as conduct a user study to determine if humans find \OurMethod{} useful in practice. 

\pagebreak

\subsubsection*{Acknowledgements}
We want to thank Li Chen as well as the anonymous reviewers for the thoughtful feedback they provided on the paper. 
This research was supported by the Netherlands Organisation for Scientific Research (NWO) under project nr. 652.001.003, the Dutch National Police, the Italian Ministry of Education, University and Research (MIUR) under the grant ``Dipartimenti di eccellenza 2018-2022'' of the Department of Computer Science and the Department of Computer Engineering at Sapienza University of Rome. 
All content represents the opinion of the authors, which is not necessarily shared or endorsed by their respective employers and/or sponsors.

\bibliography{thesis_gnn}
\bibliographystyle{plainnat}


\clearpage
\appendix

\thispagestyle{empty}

\onecolumn \makesupplementtitle

\appendix

\section{Results Table Including Standard Deviations}
Here we show Table 2 from the manuscript including standard deviations. 
We report standard deviation for two metrics: \textit{Explanation Size} and \textit{Sparsity}, since both of these involve taking the mean over the entire dataset. 
Standard deviation does not apply to \textit{Fidelity} or \textit{Accuracy} because these metrics represent a proportion as opposed to a mean.

\begin{table}[h]
\centering
\caption*{Table 2: Results comparing our method (denoted \OurShort{}) to \baserand{}, \basekeep{}, and \baserm{}. Below each metric, $\blacktriangledown$ indicates a low value is desirable, while $\blacktriangle$ indicates a high value is desirable.}
\label{table:results}
\resizebox{\textwidth}{!}{\begin{tabular}{lrrrrrrrrrrrr}
\toprule
\multicolumn{1}{c}{} & \multicolumn{4}{c}{\synfour{}}                                                                                                                 & \multicolumn{4}{c}{\synfive{}}                                                                                                                   & \multicolumn{4}{c}{\synone{}}                                                                                                                  \\ 
\cmidrule(r){2-5}\cmidrule(r){6-9}\cmidrule{10-13} 
               & \multicolumn{1}{c}{\textit{Fid.}} & \multicolumn{1}{c}{\textit{Size}} & \multicolumn{1}{c}{\textit{Spars.}} & \multicolumn{1}{c}{\textit{Acc.}} & \multicolumn{1}{c}{\textit{Fid.}} & \multicolumn{1}{c}{\textit{Size}} & \multicolumn{1}{c}{\textit{Spars.}} & \multicolumn{1}{c}{\textit{Acc.}} & \multicolumn{1}{c}{\textit{Fid.}} & \multicolumn{1}{c}{\textit{Size}} & \multicolumn{1}{c}{\textit{Spars.}} & \multicolumn{1}{c}{\textit{Acc.}} \\


Method & \multicolumn{1}{c}{$\blacktriangledown$} &\multicolumn{1}{c}{$\blacktriangledown$} &\multicolumn{1}{c}{$\blacktriangle$} & \multicolumn{1}{c}{$\blacktriangle$} & \multicolumn{1}{c}{$\blacktriangledown$} &\multicolumn{1}{c}{$\blacktriangledown$} &\multicolumn{1}{c}{$\blacktriangle$} & \multicolumn{1}{c}{$\blacktriangle$} & \multicolumn{1}{c}{$\blacktriangledown$} &\multicolumn{1}{c}{$\blacktriangledown$} &\multicolumn{1}{c}{$\blacktriangle$} & \multicolumn{1}{c}{$\blacktriangle$} \\
\midrule
\baserand{}               & \textbf{0.00}                     & 4.70   $\pm$       4.28                    & 0.79     $\pm$ 0.07                           & 0.63                               & \textbf{0.00}                     & 9.06    $\pm$ 6.81                          & 0.75      $\pm$ 0.07                          & 0.77                               & \textbf{0.00}                     & 503.31   $\pm$ 332.61                         & 0.58   $\pm$ 0.10                             & 0.17                              \\
\basekeep{}                 & 0.32                              & 15.64      $\pm$ 12.36                       & 0.13       $\pm$ 0.06                         & 0.45                               & 0.32                              & 29.30      $\pm$16.53                       & 0.09      $\pm$ 0.04                          & 0.72                               & 0.60                              & 504.18    $\pm$ 567.92                        & 0.05      $\pm$ 0.05                          & 0.18                              \\
\baserm{}              & 0.46                              & 2.11          $\pm$ 1.04                    & 0.89         $\pm$ 0.04                       & ---                                  & 0.61                              & 2.27         $\pm$ 1.28                     & 0.92      $\pm$ 0.04                          & ---                                  & 0.21                              & 10.56      $\pm$ 20.11                       & 0.97     $\pm$ 0.04                           & \textbf{0.99}                     \\


\midrule
\OurShort{}               & 0.21                              & \textbf{2.09}     $\pm$ 2.21                & \textbf{0.90}    $\pm$ 0.07                   & \textbf{0.94}                      & 0.07                              & \textbf{1.47}    $\pm$0.77                 & \textbf{0.94}    $\pm$ 0.04                   & \textbf{0.96}                      & 0.39                              & \textbf{2.39}    $\pm$ 1.39                 & \textbf{0.99}      $\pm$ 0.01                 & 0.96                 \\
\bottomrule
\end{tabular}}
\end{table}

All differences between \OurMethod{} and the baselines are statistically significant with $\alpha=0.01$ using a $t$-test, with two exceptions: \OurMethod{} vs. \baserm{} on the \synfive{} dataset, for (i) \textit{Explanation Size} and (ii) \textit{Sparsity}. However, \OurMethod{} outperforms \baserm{} significantly on \textit{Fidelity} and \textit{Accuracy} ($\alpha=0.01$).

We do not calculate standard deviations for Table 3, where we compare against \gnnexplainer{}, since we cannot evaluate \gnnexplainer{} on \textit{Explanation Size} or \textit{Sparsity}. This is because the user must specify the \textit{Explanation Size} in advance (see Sections 6.2, 7.2). We can only evaluate \gnnexplainer{} on \textit{Fidelity} and \textit{Accuracy}, neither of which require standard deviation calculations since they represent proportions.

\end{document}